\documentclass[11pt,letterpaper]{article}
\usepackage[top=0.85in,left=1.5in,footskip=0.75in,marginparwidth=2in]{geometry}

\usepackage[utf8]{inputenc}

\usepackage{cite}

\usepackage{nameref,hyperref}

\usepackage{amsmath}
\usepackage[right]{lineno}
\usepackage{microtype}
\DisableLigatures[f]{encoding = *, family = * }

\raggedright
\usepackage{changepage}

\usepackage[aboveskip=1pt,labelfont=bf,labelsep=period,singlelinecheck=off]{caption}
\makeatletter
\renewcommand{\@biblabel}[1]{\quad#1.}
\makeatother

\usepackage{lastpage,fancyhdr,graphicx}
\usepackage{epstopdf}
\fancyheadoffset[L]{2.25in}
\fancyfootoffset[L]{2.25in}

\usepackage{color}
\definecolor{Gray}{gray}{.25}

\usepackage{graphicx}
\usepackage{sidecap}

\usepackage{wrapfig}
\usepackage[pscoord]{eso-pic}
\usepackage[fulladjust]{marginnote}
\reversemarginpar

\begin{document}
\vspace*{0.35in}

\begin{flushleft}
{\Large
\textbf\newline{A novel methodology on distributed representations of proteins using their interacting ligands}
}
\newline
\\
Hakime \"{O}zt\"{u}rk\textsuperscript{1},
Elif Ozkirimli\textsuperscript{2,*},
and Arzucan \"{O}zg\"{u}r\textsuperscript{1,*}

\bigskip
\bf{1} Department of Computer Engineering, Bogazici University, Istanbul, 34342, Turkey
\\
\bf{2} Department of Chemical Engineering, Bogazici University, Istanbul, 34342, Turkey
\\
\bigskip
* arzucan.ozgur@boun.edu.tr; elif.ozkirimli@boun.edu.tr

\end{flushleft}
\justify
\section*{Abstract}
The effective representation of proteins is a crucial task that directly affects the performance of many bioinformatics problems. Related proteins usually bind to similar ligands. Chemical characteristics of  ligands are known to capture the functional and mechanistic properties of proteins suggesting that a ligand based approach can be utilized in protein representation. In this study, we propose SMILESVec, a SMILES-based method to  represent ligands  and a novel method to compute similarity of proteins by describing them based on their ligands. The proteins are defined utilizing the word-embeddings of the SMILES strings of their ligands. The performance of the proposed protein description method is evaluated in protein clustering task using TransClust and MCL algorithms. Two other protein representation methods that utilize protein sequence, BLAST and ProtVec, and two compound fingerprint based protein representation  methods are compared. We showed that ligand-based protein representation, which uses only SMILES strings of the ligands that proteins bind to, performs as well as protein-sequence based representation methods in  protein clustering. The results suggest that ligand-based protein description can be an alternative to the traditional sequence or structure based representation of proteins and this novel approach can be applied to different bioinformatics problems such as prediction of new protein-ligand interactions and protein function annotation.  

\section*{Introduction}

The aging population is putting drug design studies under pressure as we see an increase in the incidence of complex diseases. Multiple proteins from different protein families or protein networks are usually implicated in these complex diseases such as cancer, cardiovascular, immune and neurodegenerative diseases  
\cite{poornima2016network,hu2016network,santiago2014network}. Reliable representation of proteins plays a crucial role in the performance of many bioinformatics tasks such as protein family classification and clustering, prediction of protein functions and prediction of the interactions between protein-protein and protein-ligand pairs. Proteins are usually represented based on their sequences 
\cite{chou2001prediction,cai2003svm,iqbal2013distance}. A recent study adapted Word2Vec \cite{mikolov2013distributed}, which is a widely-used word-embeddings model in Natural Language Processing (NLP) tasks, into the genomic space to describe proteins as real-valued continuous vectors using their sequences, and utilized these vectors to classify proteins \cite{asgari2015continuous}.  However, even though the structure of a protein is determined by its sequence, sequence alone is usually not adequate to completely understand its mechanism. Furthermore, the relationship between fold or architecture and function was shown to be weak, while a strong correlation was reported for architecture and bound ligand   \cite{martin1998protein}. Semantic features such as functional categories and annotations, and Gene Ontology (GO) classes \cite{nascimento2016multiple,shi2015predicting,cao2016integrated,frasca2017multitask} have been suggested to support the functional understanding of proteins,  nevertheless these features are usually described in the form of binary vectors preventing the direct use of the provided information. Therefore, a novel approach that defines proteins by integrating functional characterizations can provide important information toward understanding and predicting protein structure, function and mechanism. Ligand-centric approaches are  based on the chemical similarity of compounds that interact with similar proteins \cite{peon2016reliable} and have been successfully adopted for  tasks such as target fishing, off-target effect prediction and protein-clustering \cite{chiu2014homopharma,schenone2013target}.  The use of chemical similarity of the interacting ligands of proteins to group them resulted in both biologically and functionally related  protein clusters \cite{keiser2007relating,ozturk2015classification}.  Motivated by these results, we propose to describe proteins using their interacting ligands.

In order to define the protein with a ligand centric approach, the description of the ligand is critical. Ligands can be represented in many different forms including knowledge-based fingerprints, graphs, or strings. Simplified Molecular Input Line Entry System (SMILES), which is a character-based representation of ligands,  has been  used for QSAR studies \cite{schwartz2013smifp,cao2012silico} and protein-ligand interaction prediction \cite{ozturk2016comparative,jastrzkebski2016learning}. Even though it is a string based representation form, use of SMILES performed as well as powerful graph-based representation methods in protein-ligand interaction prediction and has  proven to be computationally less expensive \cite{ozturk2016comparative}. A recent study that employed a Recurrent Neural Networks (RNN) based model to describe compound properties also used SMILES to predict chemical properties \cite{goh2017smiles2vec}. However, such deep-learning based approaches require more computational power.
An advantage of SMILES is that it provides a promising environment for the adoption of NLP approaches because it is character based. Distributed word representation models have been widely used in recent studies of NLP tasks, especially with the introduction of Word2Vec \cite{mikolov2013distributed}. The model requires a large amount of text data to learn the representations of words to describe them in low-dimensional space as real valued vectors. These vectors comprise the syntactic and semantic features of the words, e.g., the vectors of words with similar meanings are also similar. 

In this study, we introduce SMILESVec, in which we adopted the word-embeddings approach to define ligands by utilizing their SMILES strings. Ligands are represented  by learning features from a large SMILES corpus via Word2Vec \cite{mikolov2013distributed}, instead of using manually constructed ligand features as it is done in fingerprint models. We then describe each protein  using the average of its interacting ligand vectors that are built by SMILESVec. We followed a similar pipeline for evaluation that is presented in \cite{bernardes2015evaluation} in which the authors compared the performances of different clustering algorithms on the task of detecting remote homologous protein families. We measured how well SMILESVec-based protein representation  describes proteins within a protein clustering task by using two  state-of-the-art clustering algorithms; Transitive Clustering (TransClust) \cite{wittkop2010partitioning} and Markov Clustering Algorithm (MCL) \cite{enright2002efficient}. 

The performance of clustering using  SMILESVec-based protein representation was compared with that using the traditional BLAST, MACCS-based \cite{willighagen2017chemistry} and Extended Fingerprint-based protein representations as well as the recently proposed distributed protein vector representation, which is called ProtVec \cite{asgari2015continuous}.  ASTRAL data set (A-50) of SCOPe database was used as benchmark  \cite{chandonia2017scope,fox2013scope}. 

The results showed that the representation of proteins with their ligands is a promising method with competitive F-scores in the protein clustering task, even though no sequence or structure information is used. SMILESVec can be an alternative approach to binary-vector based fingerprint models for ligand-representation. The ligand-based protein representation might be useful in different bioinformatics tasks such as identifying new protein-ligand interactions and protein function annotations.

\section*{Materials and Methods}

\subsection*{Data set}
The ASTRAL data sets are the part of Structural Classification of Proteins (SCOP) collection and classified under folds, families and super-families \cite{fox2013scope}. A family denotes a group of proteins with typically distinct functionalities but also with high sequence similarities, whereas a super-family is a group of protein families with structural and functional similarities amongst families. 
The ASTRAL data sets are named based on the minimum sequence similarity of the proteins that they comprise.  For instance, ASTRAL 50 (A-50) data set includes proteins with at most 50\% sequence similarity (http://scop.berkeley.edu/astral/subsets/ver=1.75\&seqOption=1). In this study, we used A-50  data set from SCOP 1.75 version to demonstrate the performance of the protein representation methods and considered clustering into families and super-families for evaluation. Families and super-families with single protein were removed while preparing the data \cite{bernardes2015evaluation}. We used the same protein pairs that \cite{bernardes2015evaluation} used for A-50 to compute similarity scores\\ (http://www.lcqb.upmc.fr/julianab/software/cluster/). 

\subsection*{ Collection of Protein-Ligand Interactions} \label{sec:collectdata}
\begin{figure}[h]
\centerline{\includegraphics[scale=0.18]{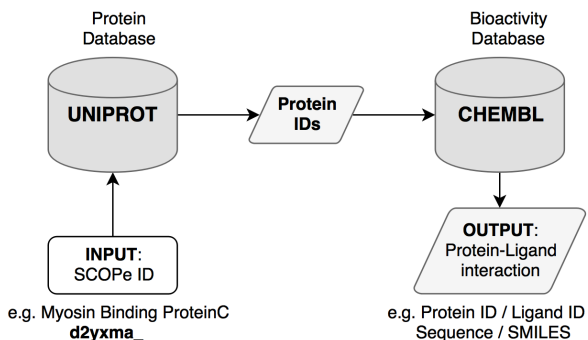}}
\caption{Extraction of protein-ligand interactions. As an example protein, Myosin Binding ProteinC is provided as input with its corresponding SCOPe ID: \textbf{d2yxma\_}}\label{fig:01}
\end{figure}

First, the corresponding UniProt identifiers were extracted for each protein in A-50 dataset using Bioservices Python package \cite{cokelaer2013bioservices}. Then, the interacting ligands with their corresponding canonical SMILES were retrieved from ChEMBL  using ChEMBL web services \cite{davies2015chembl} (Data collected on Dec 30, 2017).  The workflow of protein-ligand interaction extraction is illustrated in Figure \ref{fig:01}.  The collected interactions were used to build the proposed SMILESVec-based protein representations.

\subsection*{Distributed Representation of Proteins and Ligands}\label{section:distributed}
The Word2Vec model, which is based on feed-forward neural networks, has been previously adopted to represent proteins using their sequences \cite{asgari2015continuous}.  The approach, that we will refer to as ProtVec throughout the article, improved the  performance for the protein classification problem. In this study, we used the Word2Vec model with the Skip-gram approach to consider the order of the surrounding words.  In the biological context, we can use the string representations	 of proteins/ligands (e.g., FASTA sequence for proteins and SMILES for ligands) in textual format and define words as sub-sequences of these representations. 

\begin{figure*}[!tpb]
\centerline{\includegraphics[scale=0.18]{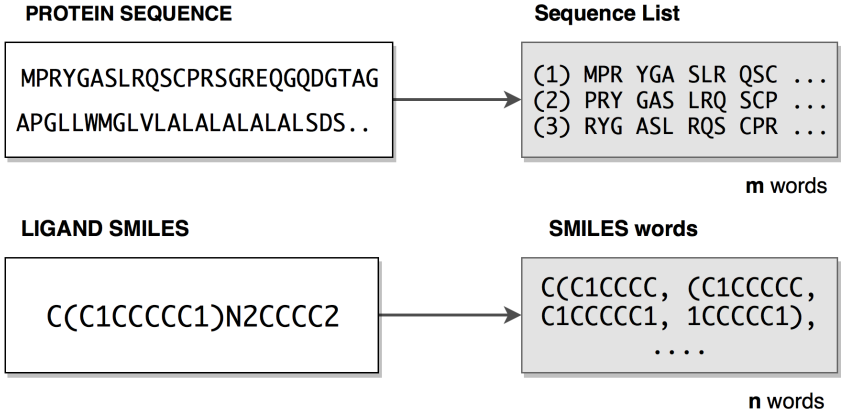}}
\caption{ Representation of biological and chemical words .}\label{fig:02}
\end{figure*}

Figure \ref{fig:02} illustrates a sample protein sequence and its sequence list (biological words) as well as a sample ligand SMILES and its corresponding sub-sequences (chemical words). The biological words which are referred to as sequence-lists are created with a set of three characters of non-overlapping sub-sequences for each list that starts from the character indices 1,2, and 3, respectively, therefore leading to three sequence lists \cite{asgari2015continuous}.  The chemical words were created as 8-character long overlapping substrings of SMILES with sliding window approach. As shown in Figure \ref{fig:02}, the SMILES string ``C(C1CCCCC1)N2CCCC2" is divided into the following chemical words: ``C(C1CCCC", ``(C1CCCCC", ``C1CCCCC1", ``1CCCCC1), ``CCCCC1)N", ``CCCC1)N2", ... , ``)N2CCCC2". We performed several experiments in which word size varied in the range of 4-12 characters and 8-charactered chemical words obtained the best results. 

With the use of the Word2Vec model we were able to describe complex structures using their simplified representations. For each subsequence (word) that was extracted from protein sequence/ligand SMILES, Word2Vec produced a real-valued vector that is learned from a large training set. The vector learning is based on the context of each subsequence (e.g. its surrounding subsequences) and can detect some important subsequences that usually occur in the same contexts. Therefore, with the help of the neural-network based nature of Word2Vec, every subsequence of a protein sequence/ligand SMILES was described in a semantically meaningful way.
The Word2Vec model defined a vector representation for each of the 3-residue subsequences of the proteins.  Protein vectors were constructed as the average of the summation of these subsequence vectors as described in Equation \ref{eq:01} where $vector(subsequence_k)$ refers to the 100-dimensional real-valued vector for the $k_{th}$ subsequence and $m$ is equal to the total number of sub-sequences that can be extracted from a protein sequence.  For proteins, 550K protein sequences from UniProt were used to train Word2Vec with skip-gram approach.
\begin{equation} \label{eq:01}
ProtVec = vector(protein) =  \frac{\sum_{k=1}^m vector(subsequence_k)}{m} 
\end{equation}

Similarly, the Word2Vec model produced a real-valued vector for each SMILES word and the corresponding ligand vector is constructed as the average of the summation of these SMILES word vectors as described in Equation \ref{eq:02}. $Vector(subsequence_k)$ represents the Word2Vec output for the 8-character long $k_{th}$ subsequence of the SMILES string and $n$ indicates the total number of these SMILES subsequences (words).  We will refer to ligand vectors as SMILESVec throughout the article.  For learning, 1.7M canonical SMILES from CHEMBL database 
(ftp.ebi.ac.uk/pub/databases/chembl/ChEMBLdb/releases/$chembl_{23}$) were retrieved  \cite{wang2016pubchem}. The skip-gram approach with vector size set to 100 was used.
\begin{equation} \label{eq:02}
SMILESVec = vector(ligand) =  \frac{\sum_{k=1}^n vector(subsequence_k)}{n} 
\end{equation}
We also used the Word2Vec model to learn embeddings for the characters in the SMILES alphabet. Therefore instead of word-level, we created char-level embeddings for the unique characters that appear in SMILES in $ChEMBL_{23}$ data set (58 chars).  Equation \ref{eq:03} describes $SMILESVec_{char}$ where $n$ in this case represents the total number of the characters in a SMILES.
\begin{equation} \label{eq:03}
SMILESVec_{char} = vector(ligand) =  \frac{\sum_{k=1}^n vector(char_k)}{n} 
\end{equation}

We further investigated an important aspect when working with SMILES representation, since  there are several valid SMILES for a single molecule. Canonicalization algorithms were coined for the purpose of generating a unique SMILES for a molecule, however couldn't prevent the diversity that came with different canonicalization algorithms. Thus, it is not that surprising that canonical SMILES definition can differ from database to database. ChEMBL uses  Accelrys's Pipeline Pilot that uses an algorithm derived from Daylight's \cite{papadatos2014chembl}, whereas Pubchem uses OpenEye software \cite{oechem2012openeye} for canonical SMILES generation \cite{balakin2009pharmaceutical}. The most evident difference between the canonical SMILES  of two databases is that ChEMBL includes isomeric information, whereas Pubchem does not. Therefore, even though we collected the SMILES of the interacting ligands from the ChEMBL database, we both experimented learning chemical words and characters from  ChEMBL and Pubchem  canonical SMILES corpora both separately and together (combined). 

We can  represent a protein/ligand vector as the output of the  maximum or minimum functions, where $m$ is the total number of the subsequences that are created from the protein/ligand sequence and $d$ is the dimensionality of the vector (i.e. the number of features). $MIN_i$ represents the minimum value of the $i^{th}$ feature among $m$ $subsequences$ (Equation \ref{eq:003}). To obtain a protein vector of minimum, $MIN_i$ is selected for each feature as defined in Equation \ref{eq:013}. Similarly, $MAX_i$ represents the maximum value of the $i^{th}$ feature among $m$ $subsequences$ (Equation \ref{eq:023}) and protein vector of maximum is created as in Equation \ref{eq:033} for $d$ number of features. The concatenation of these minimum and maximum protein vectors results in a vector with twice the dimensionality of the original vectors \cite{de2016representation}. The min/max representation is described in Equation \ref{eq:031}. 

\begin{equation} \label{eq:003}
MIN_i = min([subsequence_{0i}, subsequence_{mi}])  
\end{equation}
\begin{equation} \label{eq:013}
vector_{min}(protein) = [ MIN_0 MIN_1 ... MIN_i ... MIN_d]
\end{equation}
\begin{equation} \label{eq:023}
MAX_i = max([subsequence_{0i}, subsequence_{mi}])
\end{equation}
\begin{equation} \label{eq:033}
vector_{max}(protein) = [ MAX_0 MAX_1 ... MAX_i ... MAX_d]
\end{equation}
\begin{equation} \label{eq:031}
vector_{minmax}(protein) = [vector_{min}(protein) ]   [vector_{max}(protein)  ]  
\end{equation}

\subsection*{Protein Similarity Computation}

We used BLAST and ProtVec-based methods as baseline to compare to the ligand-centric protein representation that we proposed.
\subsubsection*{BLAST}
Basic Local Alignment Tool (BLAST) reports the similarity between protein sequences using local alignment \cite{altschul1990basic}. For the ASTRAL data sets, we used both BLAST sequence identity values and BLAST e-values that were previously obtained \cite{bernardes2015evaluation} with all-versus-all BLAST with e-value threshold of 100. 
\subsubsection*{Word Frequency-based Protein Similarity}

Word frequency-based protein similarity method uses three-charactered protein words that are created as it was explained in Section \ref{section:distributed}. However, instead of learning process, we simply count the occurrence of protein words that appear in a protein sequence. In order to compute similarity between two proteins, we used the formula depicted in Equation \ref{eq:33} \cite{vidal2005lingo}:

\begin{equation} \label{eq:33}
WordFrequency_{sim}(L_1, L_2) = \frac { \sum_{i=1}^{m}  1 -   \frac { \left| N_{P_1,i} - N_{P_2,i} \right| } { \left| N_{P_1,i} + N_{P_2,i} \right| } } {m} 
\end{equation} 

where $m$ is the total number of unique words created from protein sequences $P_1$ and $P_2$, $N_{P_1,i}$ is  the frequency of words of type $i$ in protein $P_1$ and $N_{P_2,i}$ is the frequency of words of type $i$ in protein $P_2$.

\subsubsection*{ProtVec-based Protein Similarity}
In ProtVec based clustering, protein vectors were constructed as defined in Section 2.3, either with the average or minmax method. Cosine similarity function was used to compute the similarity between two protein vectors $P1$ and $P2$ as in Equation \ref{eq:04} where $d$ denotes the size (dimensionality) of the vectors.
\begin{equation} \label{eq:04}
 CosSim(P1, P2) = \frac { \sum_{i=1}^{d} P1_i P2_i} {\lVert P1\rVert \lVert P2\rVert }
\end{equation}

\subsubsection*{SMILESVec-based Protein Similarity}
First, the ligand vectors were constructed by SMILESVec approach described in Section 2.3. Then each protein was represented as the average of the vectors of the ligands they interact with. Equation \ref{eq:05} describes the construction of a protein vector from its binding ligands where SMILESVec represents the ligand vector and $p$ represents the total number of ligands that the protein interacts with. 
\begin{equation} \label{eq:05}
vector(protein) =  \frac{\sum_{k=1}^p vector(SMILESVec_k)}{p} 
\end{equation}
Similarly, protein similarity is computed using the cosine similarity function.

\subsubsection*{Fingerprint-based Protein Similarity}
We used two popular fingerprint-based compound representation methods as an alternative to SMILESVec, namely MACCS and Extended Fingerprint.  Chemical Development Kit descriptors were used to build MACCS and Extended fingerprints of the ligands \cite{willighagen2017chemistry}. Fingerprints are binary (absence/presence) vector representations of ligands where each bit refers to  chemical features such as specific substructures and rings in which MACCS and Extended Fingerprint encode 166 and 1024 bits, respectively. The proteins were represented as described in Equation \ref{eq:06} in which  fingerprints were used to represent each interacting ligand.
\begin{equation} \label{eq:06}
vector(protein) =  \frac{\sum_{k=1}^p vector(FingerprintMethod_k)}{p} 
\end{equation}
Fingerprints were used in order to compare a knowledge-based ligand description with a data-driven approach (SMILESVec).

\subsubsection*{SMILES word frequency-based Protein Similarity}
For each interacting ligand of a protein, 8-character-long SMILES words were created as explained in Section \ref{section:distributed}. Then similarity between two proteins were computed as in Equation \ref{eq:33} using the collection of chemical words of  their respective  interacting  ligands.

\subsection*{Clustering Algorithms}
We evaluated  the effectiveness of the different protein representation approaches for the task of protein clustering. Transitivity Clustering (TransClust), which has been shown to produce the best F-measure score amongst several other algorithms in protein clustering \cite{bernardes2015evaluation} and the commonly used Markov Clustering Algorithm (MCL) were used as the protein clustering algorithms.

\subsubsection*{Transitivity Clustering (TransClust)}
TransClust is a clustering method that is based on the weighted transitive graph projection problem \cite{wittkop2010partitioning}. The main idea behind TransClust is to construct transitive graphs by adding or removing edges from an intransitive graph using a weighted cost function.  Weighted cost function is calculated as the distance between a user-defined threshold and a pairwise similarity function. 
TransClust connects two proteins on the network if their similarity is greater than the user-defined threshold. The graph is expanded by adding or removing edges until it becomes a disjoint union of cliques \cite{bernardes2015evaluation}.

\subsubsection*{Markov Clustering Algorithm (MCL)}
MCL is a network clustering algorithm that considers the weights of the edges (flows) in the network \cite{enright2002efficient} and utilized to build  a flow matrix of the network. The algorithm is implemented for a given number of iterations. The iteration number is called granularity inflation defining the homogeneity and the heterogeneity of the clusters.  We used the default value (2.0) of the inflation parameter in MCL.

\subsection*{Evaluation}
In order to evaluate the performance of the proposed methods, we utilized the F-measure, precision and recall metrics. These metrics are widely used in the evaluation of classification methods. To adapt these metrics into the assessment of clustering task, we followed the formulation explained by Bernardes and co-workers \cite{bernardes2015evaluation}. 

For a data set of $n$ proteins, let us assume $n_f$ represents the number of proteins that belong to the $f^{th}$ family or class, $n_g$ is the number of proteins that are placed in the $g^{th}$ cluster and $n_{fg}$ represents the number of proteins that belong to the $f^{th}$ family and are placed in the $g^{th}$ cluster.  Precision of cluster $g$ with respect to the $f^{th}$ family is computed as $precision_{fg} = n_{fg}/n_g$, whereas recall is defined as $recall_{fg} = n_{fg}/n_f$. Finally we can define F-measure as  in Equation \ref{eq:07}:
\begin{equation} \label{eq:07}
F-measure =  \frac{1}{n} \sum_f n_f max_g \frac{2 precision_{fg}recall_{fg}}{precision_{fg}+ recall_{fg}}
\end{equation}
$max_g$ indicates that for each family $f$, we compute precision and recall values for each corresponding $g$ cluster, and choose the maximum score. 

\section*{Results}

We evaluated the performance of five different protein similarity computation approaches in clustering of the A-50 dataset. The similarity approaches were BLAST, ProtVec, SMILESVec, MACCS, and Extended Fingerprint, the first two of which are protein sequence based similarity methods, whereas the latter three utilize the ligands to which proteins bind. We took word-frequency based protein similarity methods that use protein sequences and compound SMILES strings, respectively, as the baseline. Average (avg) and minimum/maximum (min/max) of the vectors were taken to build combined vectors for ProtVec and SMILESVec from their subsequence vectors. 

We performed our experiments on the A-50 dataset using two different clustering algorithms, TransClust and MCL. The ligand-based (SMILESVec, MACCS and Extended Fingerprint)  protein representation approaches require a protein to bind to at least one ligand in order to define a ligand-based vector for that protein. Therefore, we removed the proteins with no ligand binding information from both data sets. Table \ref{tab:01} provides a summary of A-50 data set before and after filtering. 

\begin{table}[h]
\centering
\caption{Distribution of families and super-families in A-50 data set before and after filtering}
\label{tab:01}
\begin{tabular}{|l|l|l|l|l}
\cline{1-4}
Data set & Num. Sequences & Super-families & Families &  \\ \cline{1-4}
\multicolumn{4}{|l|}{Before filtering}                &  \\ \cline{1-4}
A50      & 10816          & 1080           & 2109     &  \\ \cline{1-4}

\multicolumn{4}{|l|}{After filtering}                 &  \\ \cline{1-4}
A50      & 1639         &       425         &   652       &  \\ \cline{1-4}

\end{tabular}
\end{table}

Table \ref{tab:02} summarizes the top-10 most frequent families and super-families before and after filtering. We can observe that less than half of the frequent families and super-families remained in top-ten list such as Immunoglobulin (b.1.1) and Fibronectin type III (b.1.2) super-families and their descendants, Immunoglobulin I set (b.1.1.4) and Fibronectin type III (b.1.2.1) families, respectively. Super-families and families that weren't initially in the top-10 list  such as Protein-kinase like (d.144.1) super-family and nuclear-receptor binding domain (a.123.1) and their respective descendant families also made it among the frequent set of proteins when ligand interactions were taken into account.

In the filtered data set in which all proteins have an interacting ligand, there were 1057 proteins with fewer than 200 ligands ($64\%$ of the whole proteins) 101 of which were proteins with single ligands ($0.6\%$ of the whole proteins). There were 67 proteins with more than 10000 interacting ligands ($0.4\%$), thus increasing the average of the interacting ligands to 1791.
The protein with the highest number of interacting ligands was d2dpia2 (DNA polymerase iota), a protein involved in DNA repair \cite{jain2017mechanism} and implicated in esophageal squamous cell cancer \cite{zou2016dna} and breast cancer \cite{yang2004altered}, with   115018 ligands.

\begin{table*}[h]
\centering
\caption{Summary of the top-10 frequent families and super-families in A-50 data set of SCOPe given with the family name and the number of proteins that belong to them.}
\label{tab:02}
\scalebox{0.5}{
\begin{tabular}{|l|l|l|l|l|l|l|l|l|}
\hline
            & \multicolumn{4}{l|}{\textbf{Before filtering}}                                                                                                                                                                                      & \multicolumn{4}{l|}{\textbf{After filtering}}                                                                                                                                                                                                \\ \hline
            & \textbf{Super-family}                                                                                      & \textbf{\# prots.} & \textbf{Family}                                                                        & \textbf{\# prots.} & \textbf{Super-family}                                                                                      & \textbf{\# prots.} & \textbf{Family}                                                                                 & \textbf{\# prots.} \\ \hline
\textbf{1}  & \begin{tabular}[c]{@{}l@{}}P-loop containing nucleoside \\ triphosphate hydrolases\\ (c.37.1)\end{tabular} & 312           & \begin{tabular}[c]{@{}l@{}}Fibronectin type III\\ (b.1.2.1)\end{tabular}               & 100           & \begin{tabular}[c]{@{}l@{}}Protein kinase-like\\ (d.144.1)\end{tabular}                                    & 47            & \begin{tabular}[c]{@{}l@{}}Protein kinases, catalytic subunit \\ (d.144.1.7)\end{tabular}       & 39            \\ \hline
\textbf{2}  & \begin{tabular}[c]{@{}l@{}}NAD(P)-binding\\ Rossmann-fold domain\\ (c.2.1)\end{tabular}                    & 269           & \begin{tabular}[c]{@{}l@{}}Tyrosine-dependent oxidoreductases\\ (c.2.1.2)\end{tabular} & 86            & \begin{tabular}[c]{@{}l@{}}P-loop containing nucleoside \\ triphosphate hydrolases\\ (c.37.1)\end{tabular} & 43            & \begin{tabular}[c]{@{}l@{}}Fibronectin type III\\ (b.1.2.1)\end{tabular}                        & 28            \\ \hline
\textbf{3}  & \begin{tabular}[c]{@{}l@{}}Immunoglobulin\\ (b.1.1)\end{tabular}                                           & 174           & \begin{tabular}[c]{@{}l@{}}Canonical RNA-binding domain\\ (d.58.7.1)\end{tabular}      & 82            & \begin{tabular}[c]{@{}l@{}}Immunoglobulin\\ (b.1.1)\end{tabular}                                           & 41            & \begin{tabular}[c]{@{}l@{}}Eukaryotic proteases\\ (b.47.1.2)\end{tabular}                       & 25            \\ \hline
\textbf{4}  & \begin{tabular}[c]{@{}l@{}}"Winged helix" DNA-binding domain\\ (a.4.5)\end{tabular}                        & 157           & \begin{tabular}[c]{@{}l@{}}Immunoglobulin I set\\ (b.1.1.4)\end{tabular}               & 67            & \begin{tabular}[c]{@{}l@{}}NAD(P)-binding \\ Rossmann-fold domain\\ (c.2.1)\end{tabular}                   & 32            & \begin{tabular}[c]{@{}l@{}}EGF-type module\\ (g.3.11.1)\end{tabular}                            & 24            \\ \hline
\textbf{5}  & \begin{tabular}[c]{@{}l@{}}Thioredoxin-like\\ (c.47.1)\end{tabular}                                        & 149           & \begin{tabular}[c]{@{}l@{}}G proteins\\ (c.37.1.8)\end{tabular}                        & 64            & \begin{tabular}[c]{@{}l@{}}Trypsin-like serine proteases\\ (b.47.1)\end{tabular}                           & 31            & \begin{tabular}[c]{@{}l@{}}Immunoglobulin I set\\ (b.1.1.4)\end{tabular}                        & 23            \\ \hline
\textbf{6}  & \begin{tabular}[c]{@{}l@{}}(Trans)glycosidases\\ (c.1.8)\end{tabular}                                      & 126           & \begin{tabular}[c]{@{}l@{}}Immunoglobulin V set\\ (b.1.1.1)\end{tabular}               & 63            & \begin{tabular}[c]{@{}l@{}}Fibronectin type III\\ (b.1.2)\end{tabular}                                     & 28            & \begin{tabular}[c]{@{}l@{}}SH2 domain \\ (d.93.1.1)\end{tabular}                                & 22            \\ \hline
\textbf{7}  & \begin{tabular}[c]{@{}l@{}}Nucleic acid-binding proteins\\ (b.40.4)\end{tabular}                           & 110           & \begin{tabular}[c]{@{}l@{}}Classic zinc finger, C2H2\\ (g.37.1.1)\end{tabular}         & 60            & \begin{tabular}[c]{@{}l@{}}EGF/Laminin\\ (g.3.11)\end{tabular}                                             & 27            & \begin{tabular}[c]{@{}l@{}}Nuclear receptor \\ ligand-binding domain\\ (a.123.1.1)\end{tabular} & 18            \\ \hline
\textbf{8}  & \begin{tabular}[c]{@{}l@{}}Homeodomain-like\\ (a.4.1)\end{tabular}                                         & 102           & \begin{tabular}[c]{@{}l@{}}Phosphate binding protein-like\\ (c.94.1.1)\end{tabular}    & 56            & \begin{tabular}[c]{@{}l@{}}SH2 domain\\ (d.93.1)\end{tabular}                                              & 22            & \begin{tabular}[c]{@{}l@{}}Cyclin\\ (a.74.1.1)\end{tabular}                                     & 15            \\ \hline
\textbf{9}  & \begin{tabular}[c]{@{}l@{}}Fibronectin type III\\ (b.1.2)\end{tabular}                                     & 100           & \begin{tabular}[c]{@{}l@{}}PDZ domain\\ (b.36.1.1)\end{tabular}                        & 54            & \begin{tabular}[c]{@{}l@{}}Cysteine proteinases\\ (d.3.1)\end{tabular}                                     & 20            & \begin{tabular}[c]{@{}l@{}}Pleckstrin-homology domain \\ (b.55.1.1)\end{tabular}                & 15            \\ \hline
\textbf{10} &

\begin{tabular}[c]{@{}l@{}}S-adenosyl-L-methionine-dependent \\ methyltransferases\\ (c.66.1)\end{tabular} & 99            & \begin{tabular}[c]{@{}l@{}}N-acetyl transferase, NAT\\ (d.108.1.1)\end{tabular}        & 53            & 

\begin{tabular}[c]{@{}l@{}}Nuclear receptor \\ ligand-binding domain\\ (a.123.1)\end{tabular}              & 19            & \begin{tabular}[c]{@{}l@{}}Tyrosine-dependent oxidoreductases\\ (c.2.1.2)\end{tabular}          & 15            \\ \hline
\end{tabular}}
\end{table*}

Finally, we assessed the performance of the clustering algorithms with F-measure values for two different clustering scenarios, family and super-family clustering. TransClust requires a user-defined threshold to identify clusters, therefore in order to choose the best threshold  value, we computed the F-measure values for similarity threshold range of [0, 1] with 0.001 step-size for similarity computation methods that outputs in the range of 0-1.  For BLAST, range of [0, 100] with step-size value of 0.05 was tested for similarity threshold. We chose the similarity thresholds that gave the best F-measure for super-family and family to decide the final clusters.

Table \ref{tab:cluster} reports the F-measure values for family and super-family clustering and the number of clusters that are detected with TransClust and MCL algorithms, respectively. 

Between TransClust and MCL, TransClust produced better F-measure values in all representation methods on  A-50  data set. The results obtained by both clustering algorithms were better in family clustering than in super-family clustering, which was an expected outcome since detection of distantly related proteins is a much harder task. 

Both clustering algorithms relied on  similarity scores in order to group proteins. 
Among the protein sequence-based similarity methods, the poorest clustering performance in super-family/family (0.350/0.500)  belonged to BLAST with e-value, the baseline. Protein word frequency (0.686/0.744) obtained the best performance on the A-50 dataset in super-family and family clustering, respectively. The performances of the ProtVec Avg (0.681/0.739) and the ligand-based protein representation  methods followed the best result closely. Though, bringing in a semantic aspect with learning through Word2Vec model, ProtVec-based similarity (avg and minmax), was outperformed by the straightforward word-frequency based approach.

The results also showed than average-based combination method (ProtVec avg) was  better than min/max-based combination method (ProtVec minmax) to build a single protein vector from subsequence vectors in the protein clustering task. Since min/max-based combination method did not perform well in sequence-based protein similarity, we did not test the technique for SMILES-based protein similarity approaches. 


\begin{table*}[h]
\centering
\caption{Performances of TransClust and MCL algorithms in super-family and family clustering for all protein similarity computation methods with  F-measure values.}
\label{tab:cluster}
\scalebox{0.65}{
\begin{tabular}{llllllllll}
\hline
\multicolumn{1}{|l|}{}              & \multicolumn{1}{l|}{}     & \multicolumn{4}{l|}{Transclust}                                                       & \multicolumn{4}{l|}{MCL}                 \\ \hline
\multicolumn{1}{|l|}{}              & \multicolumn{1}{l|}{}     & \multicolumn{2}{l|}{Super-family}                                 & \multicolumn{2}{l|}{Family}                                       & \multicolumn{2}{l|}{Super-family}                                  & \multicolumn{2}{l|}{Family}                                        \\ \hline
\multicolumn{1}{|l|}{}              & \multicolumn{1}{l|}{}     & \multicolumn{1}{l|}{No.Clusters} & \multicolumn{1}{l|}{F-measure} & \multicolumn{1}{l|}{No.Clusters} & \multicolumn{1}{l|}{F-measure} & \multicolumn{1}{l|}{No. Clusters} & \multicolumn{1}{l|}{F-measure} & \multicolumn{1}{l|}{No. Clusters} & \multicolumn{1}{l|}{F-measure} \\ \hline
\multicolumn{10}{|c|}{Protein sequence based}    \\ \hline
\multicolumn{1}{|l|}{Blast (e-val)} & \multicolumn{1}{l|}{A-50} & \multicolumn{1}{l|}{1596}            & \multicolumn{1}{l|}{0.350}      & \multicolumn{1}{l|}{1636}         & \multicolumn{1}{l|}{0.500}  &
\multicolumn{1}{l|}{728}             & \multicolumn{1}{l|}{0.290}      & \multicolumn{1}{l|}{728}             & \multicolumn{1}{l|}{0.379}          \\ \hline
\multicolumn{1}{|l|}{Blast (identity)} & \multicolumn{1}{l|}{A-50} & \multicolumn{1}{l|}{606}            & \multicolumn{1}{l|}{0.595}      & \multicolumn{1}{l|}{660}         & \multicolumn{1}{l|}{0.631}  &
\multicolumn{1}{l|}{783}             & \multicolumn{1}{l|}{0.540}      & \multicolumn{1}{l|}{783}             & \multicolumn{1}{l|}{0.592}          \\ \hline

\multicolumn{1}{|l|}{Protein Word frequency}          & \multicolumn{1}{l|}{A-50} & \multicolumn{1}{l|}{708}            & \multicolumn{1}{l|}{0.686}         & \multicolumn{1}{l|}{688}         & \multicolumn{1}{l|}{0.744}    & \multicolumn{1}{l|}{411}             & \multicolumn{1}{l|}{0.590}          & \multicolumn{1}{l|}{411}             & \multicolumn{1}{l|}{0.606}          \\ \hline
\multicolumn{1}{|l|}{ProtVec Avg (word)}   & \multicolumn{1}{l|}{A-50} & 
\multicolumn{1}{l|}{655}            & \multicolumn{1}{l|}{0.681}       
& \multicolumn{1}{l|}{704}            & \multicolumn{1}{l|}{0.739}     
& \multicolumn{1}{l|}{1001}             & \multicolumn{1}{l|}{0.596}          & \multicolumn{1}{l|}{1001}             & \multicolumn{1}{l|}{0.665}          \\ \hline

\multicolumn{1}{|l|}{ProtVec Avg (char)}   & \multicolumn{1}{l|}{A-50} & 
\multicolumn{1}{l|}{707}            & \multicolumn{1}{l|}{0.674}       
& \multicolumn{1}{l|}{707}            & \multicolumn{1}{l|}{0.729}     
& \multicolumn{1}{l|}{1017}             & \multicolumn{1}{l|}{0.590}          & \multicolumn{1}{l|}{1017}             & \multicolumn{1}{l|}{0.662}          \\ \hline

\multicolumn{1}{|l|}{ProtVec MinMax (word)}   & \multicolumn{1}{l|}{A-50} & \multicolumn{1}{l|}{586}            & \multicolumn{1}{l|}{0.667}          & \multicolumn{1}{l|}{704}            & \multicolumn{1}{l|}{0.718}          & \multicolumn{1}{l|}{1014}             & \multicolumn{1}{l|}{0.590}          & \multicolumn{1}{l|}{1014}             & \multicolumn{1}{l|}{0.662}          \\ \hline

\multicolumn{10}{|c|}{Ligand based}    \\ \hline
\multicolumn{1}{|l|}{SMILES word frequency}   & \multicolumn{1}{l|}{A-50} & 
\multicolumn{1}{l|}{801}            & \multicolumn{1}{l|}{0.624}          &
\multicolumn{1}{l|}{957}            & \multicolumn{1}{l|}{0.704}          & \multicolumn{1}{l|}{312}             & \multicolumn{1}{l|}{0.470}          & \multicolumn{1}{l|}{312}             & \multicolumn{1}{l|}{0.475}          \\ \hline


\multicolumn{1}{|l|}{SMILESVec (word, chembl)}   & \multicolumn{1}{l|}{A-50} & 
\multicolumn{1}{l|}{621}            & \multicolumn{1}{l|}{0.677}          &
\multicolumn{1}{l|}{730}            & \multicolumn{1}{l|}{0.735}          & \multicolumn{1}{l|}{867}             & \multicolumn{1}{l|}{0.608}          & \multicolumn{1}{l|}{867}             & \multicolumn{1}{l|}{0.667}          \\ \hline

\multicolumn{1}{|l|}{SMILESVec (word, pubchem)}   & \multicolumn{1}{l|}{A-50} & 
\multicolumn{1}{l|}{573}            & \multicolumn{1}{l|}{0.668}          &
\multicolumn{1}{l|}{692}            & \multicolumn{1}{l|}{0.730}          & \multicolumn{1}{l|}{857}             & \multicolumn{1}{l|}{0.604}          & \multicolumn{1}{l|}{857}             & \multicolumn{1}{l|}{0.664}          \\ \hline

\multicolumn{1}{|l|}{SMILESVec (word, combined)}   & \multicolumn{1}{l|}{A-50} & 
\multicolumn{1}{l|}{617}            & \multicolumn{1}{l|}{0.675}          &
\multicolumn{1}{l|}{764}            & \multicolumn{1}{l|}{0.735}          & \multicolumn{1}{l|}{894}             & \multicolumn{1}{l|}{0.607}          & \multicolumn{1}{l|}{894}             & \multicolumn{1}{l|}{0.668}          \\ \hline


\multicolumn{1}{|l|}{SMILESVec(char, chembl)}   & \multicolumn{1}{l|}{A-50} & 
\multicolumn{1}{l|}{636}            & \multicolumn{1}{l|}{0.678}          & \multicolumn{1}{l|}{710}            & \multicolumn{1}{l|}{0.729}          & \multicolumn{1}{l|}{999}             & \multicolumn{1}{l|}{0.596}          & \multicolumn{1}{l|}{999}             & \multicolumn{1}{l|}{0.668}     
\\ \hline
\multicolumn{1}{|l|}{SMILESVec(char, pubchem)}   & \multicolumn{1}{l|}{A-50} & 
\multicolumn{1}{l|}{714}            & \multicolumn{1}{l|}{0.671}          & \multicolumn{1}{l|}{715}            & \multicolumn{1}{l|}{0.729}          & \multicolumn{1}{l|}{977}             & \multicolumn{1}{l|}{0.595}          & \multicolumn{1}{l|}{977}             & \multicolumn{1}{l|}{0.667}     
\\ \hline

\multicolumn{1}{|l|}{SMILESVec(char, combined)}   & \multicolumn{1}{l|}{A-50} & 
\multicolumn{1}{l|}{712}            & \multicolumn{1}{l|}{0.675}          & \multicolumn{1}{l|}{712}            & \multicolumn{1}{l|}{0.739}          & \multicolumn{1}{l|}{1006}             & \multicolumn{1}{l|}{0.595}          & \multicolumn{1}{l|}{1006}             & \multicolumn{1}{l|}{0.669}     
\\ \hline

\multicolumn{1}{|l|}{MACCS}   & \multicolumn{1}{l|}{A-50} & 
\multicolumn{1}{l|}{589}            & \multicolumn{1}{l|}{0.679}          & \multicolumn{1}{l|}{683}            & \multicolumn{1}{l|}{0.736}          & \multicolumn{1}{l|}{874}             & \multicolumn{1}{l|}{0.606}          & \multicolumn{1}{l|}{874}             & \multicolumn{1}{l|}{0.667}          \\ \hline

\multicolumn{1}{|l|}{Extended Fingerprint}   & \multicolumn{1}{l|}{A-50} & 
\multicolumn{1}{l|}{607}            & \multicolumn{1}{l|}{0.680}          & \multicolumn{1}{l|}{756}            & \multicolumn{1}{l|}{0.732}          & \multicolumn{1}{l|}{744}             & \multicolumn{1}{l|}{0.609}          & \multicolumn{1}{l|}{744}             & \multicolumn{1}{l|}{0.655}          \\ \hline

                                    &                           &                                  &                                &                                  &                                &                                   &                                &                                   &                               
\end{tabular}}
\end{table*}

Among the ligand based representation methods, we examined the performance of the word-based embeddings and character-based embeddings as well as the effect of the source of the training data set on embeddings. We collected  canonical SMILES from both ChEMBL ($~$1.7M) and Pubchem ($~$2.3M) databases. The SMILES strings of the interacting ligands were only collected from ChEMBL as explained in Section \ref{sec:collectdata}. The main difference between these two databases is that ChEMBL allows  the isometric information of the molecule to be encoded within SMILES. The results clearly indicated that the choice of the training set for embedding learning is important where SMILES was concerned. In our case, since SMILES of the interacting ligands of A-50 data set was collected from ChEMBL database, the performance of the SMILESVec in which embeddings were learned from training with ChEMBL SMILES rather than Pubchem SMILES was notably better.

We also investigated whether using the combination of the SMILES corpus of ChEMBL and Pubchem can improve the performance of SMILESVec embeddings. We indeed reported an improvement on character-based embedding in family clustering (0.739) whereas word-based embedding produced F-measure values higher than the Pubchem-based learning and lower than the ChEMBL-based learning. We can suggest that the increase in the performance of the character-based learning with the combination of two different SMILES corpora might be positively correlated with the increase in SMILES samples, while the  number of unique letters that  appear in the SMILES did not  significantly change between databases (e.g. absence/presence of the  few characters that represent isometry information). However, with the word-based learning, we  observed that there was significant increase in the variety of the chemical words, thus the combined SMILES corpus model did not work as well as it did in character-based learning. This result suggests that the size of the learning corpus may affect the representation of the embeddings, thus we might suggest a larger SMILES corpus could lead to better character-based embeddings for SMILESVec. 

Considering only ChEMBL trained SMILESVec, we observe that even though producing comparable scores, word-based approach was better than character-based SMILESVec in terms of F-measure in family clustering. In super-family clustering however, character-based approach performs as well as word-based SMILESVec. Similarly, ProtVec is also better represented in word-level rather than character-level.

The ligand-based protein representation methods, SMILESVec and MACCS-based approach performed almost as well as ProtVec in family and super-family clustering with TransClust algorithm, even though no protein sequence information was used. With MCL, a lower clustering performance was obtained compared to TransClust, and both SMILESVec  and MACCS-based method produced slightly better F-measure than  ProtVec Avg in both super-family and family clustering.  We can suggest that, since ligand-based protein representation methods capture indirect function information through ligand binding, they were recognizably better at detecting super-families than families compared to sequence-based ProtVec on a relatively distant data set.  Furthermore, SMILESVec, a text-based unsupervised learning model, produced comparable F-measure scores to MACCS and Extended fingerprints, which are binary vectors based on human-engineered feature descriptions.

Table \ref{tab:pearsonprot} reports the Pearson correlations \cite{Pearson}  among the protein similarity computation methods. Comparison with BLAST e-value resulted in negative correlation, as expected, since e-values closer to zero indicate high match (similarity). Ligand based protein representation methods had higher correlation values with BLAST e-value than protein-sequence based methods. We also observed strong correlation among the ligand-based protein representation methods, suggesting that, regardless of the ligand representation approach, the use of interacting ligands to represent proteins provides similar information.

\begin{table}[h]
\centering
\caption{Pearson correlation between protein  similarity methods}
\label{tab:pearsonprot}
\scalebox{0.8}{
\begin{tabular}{lll}
\hline
\multicolumn{1}{|l|}{Method}            & \multicolumn{1}{l|}{Method} & \multicolumn{1}{l|}{Pearson correlation} \\ \hline
\multicolumn{1}{|l|}{BLAST (e-value)}  & \multicolumn{1}{l|}{BLAST (identity)}  & \multicolumn{1}{l|}{-0.109}               \\ \hline
\multicolumn{1}{|l|}{BLAST (e-value)}  & \multicolumn{1}{l|}{Protein word frequency}  & \multicolumn{1}{l|}{-0.250}               \\ \hline
\multicolumn{1}{|l|}{BLAST (e-value)}  & \multicolumn{1}{l|}{ProtVec (avg)}  & \multicolumn{1}{l|}{-0.291}               \\ \hline
\multicolumn{1}{|l|}{BLAST (e-value)}  & \multicolumn{1}{l|}{SMILESVec (word, chembl)}  & \multicolumn{1}{l|}{-0.335}               \\ \hline
\multicolumn{1}{|l|}{BLAST (e-value)}  & \multicolumn{1}{l|}{SMILESVec (char, chembl)}  & \multicolumn{1}{l|}{-0.207}               \\ \hline
\multicolumn{1}{|l|}{BLAST (e-value)}  & \multicolumn{1}{l|}{MACCS}  & \multicolumn{1}{l|}{-0.336}               \\ \hline

\multicolumn{1}{|l|}{SMILESVec (word, chembl)}  & \multicolumn{1}{l|}{MACCS}  & \multicolumn{1}{l|}{0.895}               \\ \hline
\multicolumn{1}{|l|}{SMILESVec (char, pubchem)}  & \multicolumn{1}{l|}{MACCS}  & \multicolumn{1}{l|}{0.590}               \\ \hline
\multicolumn{1}{|l|}{SMILESVec (word, chembl)}  & \multicolumn{1}{l|}{SMILESVec (char, pubchem)}  & \multicolumn{1}{l|}{0.682}               \\ \hline 

\multicolumn{1}{|l|}{SMILESVec (word, chembl)}  & \multicolumn{1}{l|}{Extended Fp.}  & \multicolumn{1}{l|}{0.938}               \\ \hline 

\multicolumn{1}{|l|}{Extended Fp.}  & \multicolumn{1}{l|}{MACCS}  & \multicolumn{1}{l|}{0.937}               \\ \hline

\end{tabular}}
\end{table}

We further investigated a case in which similar super-family clusters were produced with SMILESVec-based protein similarity and ProtVec protein similarity using TransClust algorithm. We observed that Fibronectin Type III proteins (7 proteins) were clustered together when SMILESVec was used, whereas using ProtVec placed them into four different clusters; one cluster contained four of those proteins, another cluster contained a single protein and the other two proteins were part of other clusters. The protein that was clustered by itself (SCOPe ID:d1n26a3, Human Interleukin-6 Receptor alpha chain) had two interacting ligands (CHEMBL81;Raloxifene and CHEMBL46740;Bazedoxifene) that were also shared by a protein (SCOPe ID:d1bqua2,Cytokine-binding region of GP130) clustered separately with ProtVec. Thus, we can suggest that using information on common interacting ligands, SMILESVec achieved to combine these seven proteins into a single cluster, while ProtVec failed to do so with a sequence-based approach.

\section*{Discussion}

In this study, we first propose a ligand-representation method, SMILESVec, which uses a word embeddings model. Then, we represent proteins using their interacting ligands. In this approach, the interacting ligands of each protein in the data set are collected. Then, the SMILES string of each ligand is divided into fixed-length overlapping substrings. These created substrings are then used to build real-valued vectors with the Word2vec model and then the vectors are combined into a single vector to represent the whole SMILES string. Finally, protein vectors are constructed by taking the average of the vectors of their ligands.  The effectiveness of the proposed method in describing the proteins was measured by performing clustering on  ASTRAL 50 (A-50) dataset from the SCOPe database using two different clustering algorithms, TransClust and MCL. Both of these clustering algorithms use protein similarity scores to identify cliques. SMILESVec based protein representation was compared with other protein representation methods, namely BLAST and ProtVec, both of which depend on protein sequence to measure protein similarity, and the MACCS and Extended Fingerprint binary fingerprint based ligand-centric protein representation approaches.  The performance of the clustering algorithms, as reported by F-measure, showed that protein word-frequency based similarity model was a better alternative to BLAST e-value or sequence identity to measure protein similarity. Furthermore, ligand-based protein representation methods also produced comparable F-measure scores to ProtVec.

Using SMILESVec, we were able to define proteins based on their interacting ligands even in the absence of sequence or structure information. SMILESVec-based protein representation had better clustering performance than BLAST and comparable clustering performance to protein word-frequency based method, both of which use protein sequences. We should emphasize that SCOPe data sets were constructed based on protein similarity, thus high performance with the protein sequence-based models in family/super-family clustering is no surprise. However, having ligand-based protein representation methods, either learning from SMILES or represented with binary compound features, performing as well as protein sequence-based models is quite intriguing and promising.

SMILESVec and MACCS representation performed similarly in the task of protein clustering and better than Extended Fingerprint representation, suggesting that the word-embeddings approach that learns representations from a large SMILES corpus in an unsupervised manner is as accurate as a knowledge-based fingerprint model.  We   propose that the ligand-based representation of proteins might reveal important clues especially in protein-ligand interaction related  tasks like drug specificity or identification of proteins for drug targeting. The similarity between a candidate ligand and the SMILESVec for a protein can be used as an indicator for a possible interaction. 

We would like to mention that ASTRAL data sets contain domains rather than full length proteins while CHEMBL collects protein - ligand interaction information based on the whole protein sequence from UniProt. A multidomain protein may have multiple and diverse chemotypes of ligands binding to each domain and retrieving ligand information based on the full length protein may lump this disparate information together, leading to loss of information on domain specific ligand interactions. The performance of domain sequence based methods is therefore at an advantage because family/superfamily assignment in SCOPe is also based on domain sequence while the ligand based approach we use in SMILESVec uses more noisy data. Despite this disadvantage, ligand based approach performs as well as the sequence based approaches. We hypothesize that if domain - ligand interactions are taken into account, ligand based approaches would have higher performance.  

The study we conducted here also showed that SMILES description is sensitive to the database definition conventions, therefore SMILES strings requires careful consideration. Since we collected the protein-ligand interaction and ligand SMILES information from ChEMBL database to represent proteins, building SMILESVec vectors from the chemical words trained in ChEMBL SMILES corpus yielded better F-measure than the model in which the Pubchem SMILES corpus was used for training of the chemical words.

We showed that ligand-centric protein representation performed at least as well as protein sequence based representations in the clustering task even in the absence of sequence information. Ligand-centric protein representation is only available for proteins with at least one known ligand interaction, while a sequence based approach can miss key functional/mechanistic properties of the protein. The orthogonal information that can be obtained from the two approaches has been previously observed \cite{o2016ligand}. As future work, we will investigate combining both sequence and ligand information in protein representation. We believe that this approach will provide a deeper understanding of protein function and mechanism  toward the use of these representations in clustering and other bioinformatics tasks such as function annotation and prediction of novel protein - drug interactions.

\section*{Acknowledgments}
TUBITAK-BIDEB 2211-E Scholarship Program (to HO) and BAGEP Award of the Science Academy (to AO) are gratefully acknowledged.  We thank Prof. Kutlu O. Ulgen and Mehmet Aziz Yirik for helpful discussions.

\section*{Funding}

This work is funded by Bogazici University Research Fund (BAP) Grant Number 12304.

\nolinenumbers

\bibliography{library}

\bibliographystyle{abbrv}

\end{document}